\documentclass{article}

\usepackage{arxiv}

\usepackage[utf8]{inputenc} % allow utf-8 input
\usepackage[T1]{fontenc}    % use 8-bit T1 fonts
\usepackage{hyperref}       % hyperlinks
\usepackage{url}            % simple URL typesetting
\usepackage{booktabs}       % professional-quality tables
\usepackage{amsfonts}       % blackboard math symbols
\usepackage{nicefrac}       % compact symbols for 1/2, etc.
\usepackage{microtype}      % microtypography
\usepackage{lipsum}		% Can be removed after putting your text content
\usepackage{graphicx}
\usepackage{natbib}
\usepackage{doi}
\usepackage{bm}
	
\usepackage[varvw]{newtxmath}       % selects Times Roman as basic font
\usepackage{amsmath}
\mathchardef\mhyphen="2D % Define a "math hyphen"
\usepackage{booktabs}
\usepackage{multirow}
\usepackage{comment}

\usepackage{subcaption}
\usepackage{adjustbox}
\usepackage{threeparttable}
\usepackage{listings}

\title{Deep Neural Operator Driven Real Time Inference for Nuclear Systems to Enable Digital Twin Solutions}

\author{ 	{Kazuma ~Kobayashi} \\
	Nuclear, Plasma \& Radiological Engineering\\
	University of Illinois at Urbana-Champaign\\
	Urnaba, IL 61801, USA \\
	\And
      {Syed Bahauddin ~Alam} \\
	Nuclear, Plasma \& Radiological Engineering\\
	University of Illinois at Urbana-Champaign\\
	Urnaba, IL 61801, USA \\
 }

\renewcommand{\shorttitle}{\textit{Deep Operator Networks for Nuclear Systems} }

\hypersetup{
pdftitle={A template for the arxiv style},
pdfsubject={q-bio.NC, q-bio.QM},
pdfauthor={David S.~Hippocampus, Elias D.~Striatum},
pdfkeywords={First keyword, Second keyword, More},
}

\begin{document}
\maketitle

\begin{abstract}
This paper focuses on the feasibility of Deep Neural Operator (DeepONet) as a robust surrogate modeling method within the context of digital twin (DT)  for nuclear energy systems.  
Through benchmarking and evaluation, this study showcases the generalizability and computational efficiency of DeepONet in solving a challenging particle transport problem. DeepONet also exhibits remarkable prediction accuracy and speed, outperforming traditional ML methods, making it a suitable algorithm for real-time DT inference. 
However, the application of DeepONet also reveals challenges related to optimal sensor placement and model evaluation, critical aspects of real-world implementation. Addressing these challenges will further enhance the method's practicality and reliability.
Overall, DeepONet presents a promising and transformative nuclear engineering research and applications tool. Its accurate prediction and computational efficiency capabilities can revolutionize DT systems, advancing nuclear engineering research. This study marks an important step towards harnessing the power of surrogate modeling techniques in critical engineering domains.
\end{abstract}

% keywords can be removed
%\keywords{Explainable AI \and Interpretable AI  \and Digital Twin \and Prognostics and Health Management}
{ \small
\section{Introduction}
\label{sec:intro}

A reliable and sustainable energy supply is essential to support and drive economic activity in the modern world. As the urgent need for carbon-neutral solutions becomes increasingly evident, nuclear energy is projected to assume a significant role. With the ongoing global increase in energy demands, nuclear power is poised to emerge as a critical and environmentally friendly alternative to conventional energy sources.

Furthermore, nuclear research actively explores novel technologies and approaches to advance the field. Among these exciting research areas, using digital twin (DT) technology in nuclear systems has gained substantial attention. The United States Nuclear Regulatory Commission (U.S. NRC) recognizes the potential of DT technology and has identified it as a key area for future research \cite{yadav2021technical}. The NRC has highlighted several potential benefits of DTs in nuclear energy applications, including increased operational efficiencies, enhanced safety and reliability, reduced errors, faster information sharing, and improved predictive capabilities \cite{yadav2021technical, kobayashi2022non}. However, it is imperative to recognize that the evolution of DT technology within the realm of nuclear systems is still at its inception, bringing forth a range of intricate challenges that necessitate diligent resolution and strategic overcoming \cite{kazuma2023components}. These challenges span multifarious domains, encapsulating crucial aspects like the integrating data from various sources, the modeling \& and simulation (M\&S) of complex nuclear systems, the synchronization in real-time between the digital replica and physical asset, and the critical domains of cybersecurity and safeguarding data privacy. Furthermore, developing and deploying advanced sensors and network architecture is paramount, as they are pivotal to ensuring an uninterrupted data flow \cite{yadav2021technical, kazuma2023components}. Among these challenges, this study focuses on potentially utilizing deep learning methods in M\&S for nuclear systems.

%%%%%%%%%%%
A digital replica is created by capturing and integrating various data sources, such as sensor data, operational data, and design specifications, to generate a highly detailed and accurate representation of the physical system \cite{ibmwebsite}. This virtual replica serves multiple purposes: monitoring, analysis, simulation, and prediction. While traditional analysis and simulation codes can fulfill these roles, it can be challenging to balance accuracy and speed, especially in systems requiring rapid analysis and response prediction. Dedicated codes are employed for analysis in certain domains like nuclear reactor systems. However, these codes often prioritize high analysis accuracy, which can result in high computational costs. Relying solely on expensive simulations based on real-time information from various sensors installed in the reactor system and then making control decisions based on the results can lead to sluggish operator response times and, in extreme cases, potentially catastrophic nuclear accidents. This situation calls for a new approach in DT technology for reactor systems \cite{rahman2022leveraging, daniell2022physics, daniell2023transfer} that simultaneously addresses the need for high calculation accuracy and speed.

Deep learning, particularly through neural networks (NNs), has transformed the landscape of modeling techniques for nonlinear systems. Its significant contribution lies in its remarkable ability to capture intricate and complex relationships within data, making it exceptionally well-suited for approximating nonlinear phenomena. Within nonlinear systems, NNs excel at approximating functions that map inputs to outputs without imposing explicit assumptions about underlying dynamics. It is particularly advantageous when traditional modeling approaches falter due to the system's nonlinear, dynamic, or stochastic nature. The applications of NNs within the nuclear field further highlight their versatility. Instances include leveraging Deep Neural Networks (DNNs) {\cite{MOLOKO2023109813}} and Recurrent Neural Networks (RNNs) {\cite{CADINI2007483}} for predicting neutron flux distribution in reactor cores. Additionally, NNs have been employed for surrogate modeling of nuclear reactor dynamic equations. It has been facilitated through the use of Convolutional Neural Networks (CNNs) {\cite{hadad2007application}} and Physics-Informed Neural Networks (PINNs) {\cite{WANG2022109234}}.

The advancement of NNs has been instrumental in shaping the landscape of data-driven modeling techniques, including hybrid approaches like PINNs. However, as these deep learning models progress to the deployment phase, they encounter a formidable challenge termed "dataset shift." This challenge is rooted in the dynamic nature of real-world data. While these models are trained on specific datasets, they might only partially encompass the diverse scenarios they will face during practical applications. Environmental changes, shifts in user behavior, or other external factors can introduce variations in data distribution that the model has yet to encounter during its training. This discrepancy between the distributions of training and deployment data can lead to a decline in model performance, resulting in less reliable predictions and potentially compromising the system's overall functionality. The model might need help to adapt to novel situations insufficiently represented in its training data.

To illustrate, consider the behavior of neutrons within a nuclear system. This behavior is pivotal in reactor core analysis, shielding design, and criticality assessments. It can be effectively described by the neutron transport equation {\cite{lamarsh2001introduction}}, featuring a source term that encompasses contributions from processes like radioactive decay and fission. This neutron source term is far from constant in practical scenarios, exhibiting varying energies and spatial distributions over time. Just as the neutron source term evolves, the dataset shift phenomenon reflects the evolving nature of real-world data, creating a parallel challenge for deploying neural network models effectively in intricate systems such as nuclear reactors.

In order to address these challenges, this study introduces a potential solution in the form of data-driven surrogate modeling, utilizing the Deep Operator Network to fulfill these demanding prerequisites effectively. It is designed to handle functions as inputs and then construct an operator within a system using training data. Unlike conventional machine learning methods that usually deal with input-output mappings, DeepONet focuses on the mapping between functions {\cite{lu2021learning}}. Utilizing DeepONet's unique attribute, the study constructs a surrogate model for the intricate spatial distribution of neutron flux. This approach employs the neutron source's spatial distribution as an input function. This utilization of function-based modeling demonstrates the potential of DeepONet in capturing the intricate dynamics of nuclear systems {\cite{lu2021learning}} as a faster surrogate model instead of conventional nuclear code.

It is worth addressing that this research constitutes an extension of earlier published investigations by the lead author {\cite{kobayashi2023neural,kobayashi2023psam}}, in {\cite{kobayashi2023neural}}, a comprehensive exposition of the Digital Twin (DT) framework for real-time monitoring of the degradation of nuclear system structures, systems, and components (SSC) is presented, including illustrative instances of DeepONet applications. Moreover, {\cite{kobayashi2023psam}} encapsulates the current study's foundational concept and preliminary outcomes. This paper builds upon and expands the concepts and findings of those publications. In this extended version, we delve deeper into the practical application of the Deep Operator Network in nuclear engineering problems, providing additional insights and discussions. The current work also introduces new experimental data and presents further validations to reinforce the robustness of the proposed approach.

%The scope of this current study is only limited to the ......

\section{Deep operator network (DeepONet)}
\label{sec:deeponet}
DeepONet, an advanced neural network, is built upon the Universal Approximation Theorem but focuses on the Universal Approximation Theorem for Operators \cite{lu2021learning, chen1995universal}. While traditional neural networks map inputs to a function space, DeepONet is designed to map information from functions to an operator that can be observed in any domain.

Suppose $G$ is an operator that takes an input function $u$; this assumption makes $G(u)$ the output function corresponding to the input one. As explained in \cite{lu2021learning}, in the domain of $G(u)$, the real number at any sampling point $y$ can be expressed as $G(u)(y)$. To handle an input function in calculations, it must be discretized in the input space $V$. In the concept of DeepONet \cite{lu2021learning}, input functions are discretized by sampling at the fixed positions $\{x_{1}, x_{2}, \dots, x_{m}\}$ where $m$ represents the number of discretized points. Therefore, DeepONet can handle the two network inputs; $[u(x_1), u(x_2), \dots, u(x_m) ]^{T}$ and $y \in \mathbb{R}^{d}$. Based on the Universal Approximation Theorem for the Operator, the operator $G$ can be expressed by the Generalized Universal Approximation Theorem for the Operator by \cite{lu2021learning, CAI2021110296} as follows:

%% Theorem 2 (Generalized Universal Approximation Theorem for Operator)
\begin{equation}
\label{UAT_for_operator}
    \left| G(u)(y) - \langle \underbrace{\rm{\bf{g}}\it{\left( u(x_{1}, u_{2}, \cdots, u(x_{m}) \right)}}_{\text{branch}}, \underbrace{\rm{\bf{f}}\it{(y)}}_{\text{trunk}} \rangle  \right| < \epsilon
\end{equation}
where $\epsilon > 0$, $\bf{g}$ and $\bf{f}$ are continuous vector functions $\bf{g}: \mathbb{R}^{m} \longrightarrow \mathbb{R}^{p}$ and $\bf{f}: \mathbb{R}^{d} \longrightarrow \mathbb{R}^{p}$, $ \langle {\cdot,\cdot} \rangle$ represents the dot product in $\mathbb{R}^{p}$. The function $\bf{g}$ and $\bf{f}$ are replaced with neural networks. Here, the NNs that handle the input function are distinguished as "branch" and those that handle the input vector $y \in \mathbb{R}^{d}$ as "trunk." Since the main topic of this study is the applied use of DeepONet, please refer to \cite{lu2021learning, lu2021deepxde} for more detailed mathematical proof.

DeepONet employs the branch-trunk architecture, where the "branch" and "trunk" networks are subnetworks implemented using NNs \cite{lu2021learning, GOSWAMI2022114587}. The trunk network deals with domain information, while the branch network encodes sensor information from the function. Internal filtering allows DeepONet to handle data efficiently. This architecture is harnessed to approximate the system's operator, and the entire model is trained using loss associated with the predictions. Figure \ref{fig:deeponet} illustrates the branch-trunk architecture. For a more in-depth mathematical proof, refer to \cite{lu2021learning, lu2021deepxde}.

DeepONet is proven to be versatile in diverse domains, including multiphysics scenarios such as electric convection \cite{CAI2021110296}, bubble growth dynamics \cite{lin2021operator}, and fluid dynamics \cite{MAO2021110698}. As DeepONet continues to prove its efficacy in handling complex nonlinear and computationally intensive problems, its applicability is anticipated to expand across a wide array of fields in the future.

\begin{figure}[!htbp]
    \centering
    \includegraphics[width=\textwidth]{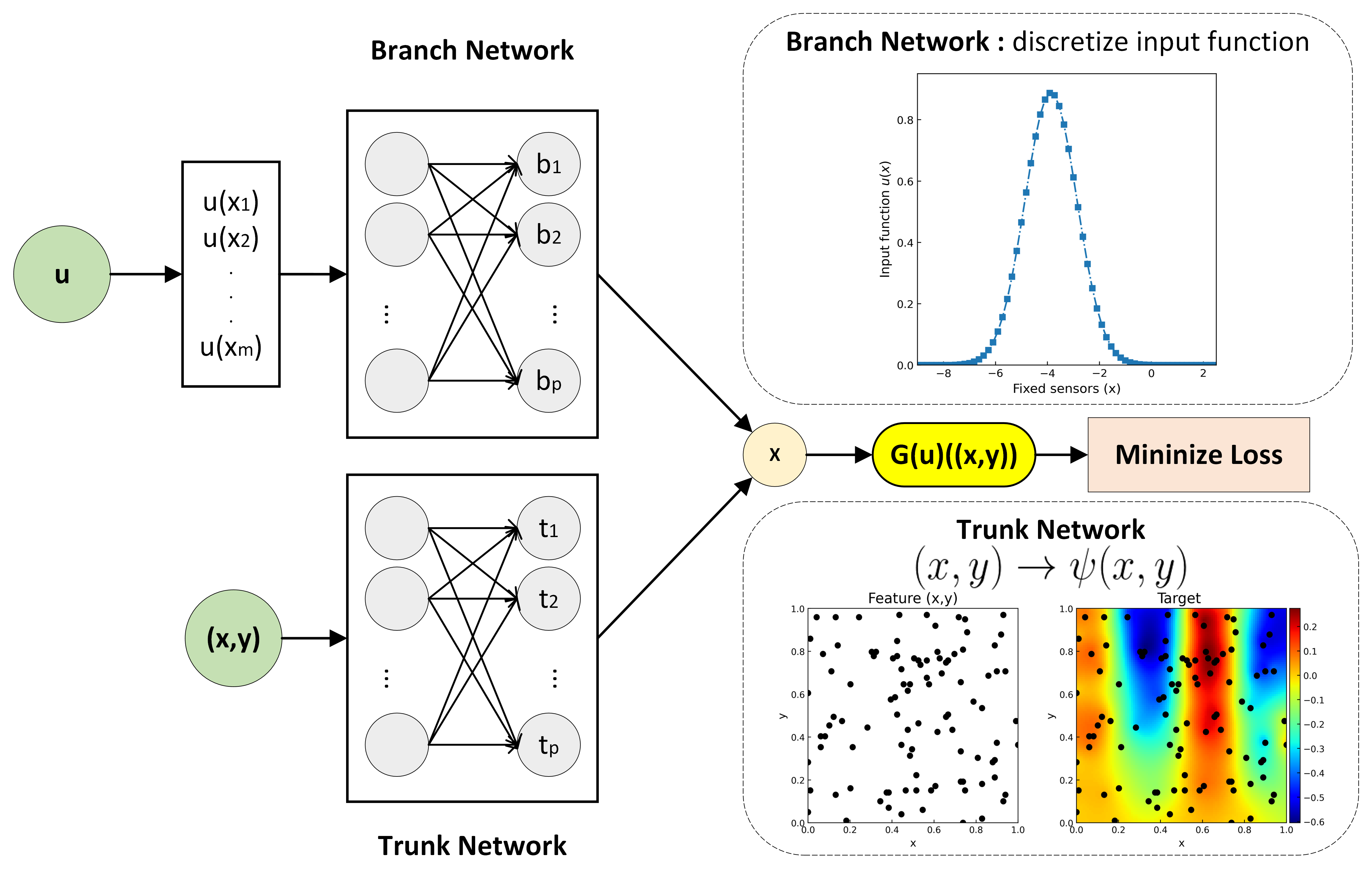}
    \caption{Illustration of DeepONet Branch-Trunk architecture employing fully-connected neural networks. It is an example when the input vector is composed of 2-dimension (i.e., $\bf{y} = $[x,y]). The discretized input functions and input vector $y$ are individually fed into fully connected neural networks. Then, the dot product is computed using their network outputs. Note: This figure was completely redrawn by the authors based on the concept of DeepONet presented in \cite{lu2021learning} to fit the problem setting of this study.}
    \label{fig:deeponet}
\end{figure}

%%%%%%%%%%%%%%%%%%%%%%%%%%%%%%%%%%%%%%%%%%
\section{Experiments}
In order to showcase the capabilities of DeepONet, a surrogate model is constructed for calculating the 2-dimensional spatial distribution of neutron flux in a maze. The training and test datasets used for training the DeepONet model are prepared using Particle and Heavy Ion Transport code System (PHITS) version 3.24 \cite{sato2018features}. This section elaborates on the methodology employed for data generation and the simulation setup utilized in this study.

\subsection{Particle Transport Code}
PHITS, a Monte Carlo particle transport simulation code, has the capability to accurately simulate particle transport across a wide range of energy spectra through the utilization of sophisticated nuclear reaction models and comprehensive data libraries \cite{sato2018features}. Its versatility allows for its application in various research fields, including accelerator technology, radiotherapy, space radiation, and other domains involving particle and heavy ion transport phenomena. In this study, we leverage the power of PHITS to simulate the 2D spatial distribution of neutron flux within a maze. To accomplish this, we have customized and employed the sample code provided by the PHITS Development Group as a foundation for our research. For particle transport calculations in PHITS, the definition of geometry, material, radiation sources, and the phenomenon to be analyzed, referred to as Tally, is necessitated. 

\subsubsection{Geometry}
In our simulation, a hypothetical maze with concrete walls enclosing an air-filled interior is chosen as the geometry, as depicted in Figure \ref{fig:maze_input_function} (a). During the configuration of the geometry, there is no need to specify the spatial resolution, such as the grid size, as it is a critical factor affecting the accuracy of the calculation and is determined when defining the Tally, as elucidated later. 

\subsubsection{Material}
The material densities of concrete and air are set to $2.2$ $\rm g/cm^3$ and  $1.2\times 10^{-3}$ $\rm g/cm^3$, respectively. For each defined element, nuclear reaction cross-sections are bundled from the nuclear data library JENDL-4.0 \cite{shibata2011jendl}. Moreover, the detail of material compositions used in this study is provided in Supplementary Material A. %\ref{sec:material_table}. 

\subsubsection{Neutron Source}
\label{sec:source}
The simulation employs a Gaussian distribution neutron source, where the spatial distribution of neutrons is modeled as independent Gaussian distributions along each component of the spatial vector \(\mathbf{x} = [x_{1}, x_{2}, x_{3}]\). Given the independence of the variables, the covariance matrix \(\Sigma\) is a diagonal matrix with variances \(\sigma_{j}^{2}\) for each spatial dimension, as illustrated in Equation \ref{eq:cov_matric}:

\begin{equation}
\Sigma =
\begin{pmatrix}
\sigma_{1}^{2} & 0 & 0 \\
0 & \sigma_{2}^{2} & 0 \\
0 & 0 & \sigma_{3}^{2}
\end{pmatrix}.
\label{eq:cov_matric}
\end{equation}
In this representation, the subscript \( j \) corresponds to each spatial direction. Consequently, the 3-dimensional Gaussian distribution \(\phi(\mathbf{x})\) is defined by Equation \ref{eq:multi_var_normal_distribution}:

\begin{equation}
\phi(\mathbf{x}) = \prod_{j=1}^{3} \frac{1}{\sqrt{2\pi\sigma_{j}^{2}}} \exp \left\{-\frac{1}{2 \sigma_{j}^{2}} \left(x_{j} - \mu_{j}\right)^2 \right\}.
\label{eq:multi_var_normal_distribution}
\end{equation}
The neutron source \( u(E, \mathbf{x}) \) is characterized by a mono-energetic distribution, where all neutrons possess the same energy level \( E \). This leads to the introduction of a new quantity, \(\phi(\mathbf{x}) \cdot E\), which represents the energy density distribution of neutrons in space. The neutron source distribution is thus simplified to:

\begin{equation}
u(E, \mathbf{x}) = \phi(\mathbf{x}) \cdot E = E \cdot \prod_{j=1}^{3} \frac{1}{\sqrt{2\pi\sigma_{j}^{2}}} \exp \left\{-\frac{1}{2 \sigma_{j}^{2}} \left(x_{j} - \mu_{j}\right)^2 \right\},
\label{eq:neutron_source}
\end{equation}
In this equation, \(\phi(\mathbf{x})\) represents the spatial distribution of neutron intensity and \( E \) is the constant energy level of each neutron. The product \(\phi(\mathbf{x}) \cdot E\) thus describes the distribution of neutron energy density across the spatial dimensions, providing a comprehensive view of both the spatial and energetic characteristics of the neutron source. In this equation, \( \mu_{j} \) represents the mean position component of the mean vector \( \mathbf{\mu} \). This model allows for the simulation of the spatial and energetic distribution of neutrons, providing insights into their behavior in a controlled environment.

\begin{comment}
The simulation utilizes a Gaussian distribution neutron source. This source is characterized by independent Gaussian distributions along each component of the vector $\mathbf{x} = [x_{1}, x_{2}, x_{3}]$. Assuming independent variables, the covariance matrix $\Sigma$ takes the form of a diagonal matrix with variances $\sigma_{j}^{2}$ for each variable, as shown in Equation \ref{eq:cov_matric}:

\begin{equation}
\Sigma =
\begin{pmatrix}
\sigma_{1}^{2} & 0 & 0 \\
0 & \sigma_{2}^{2} & 0 \\
0 & 0 & \sigma_{3}^{2}
\end{pmatrix}.
\label{eq:cov_matric}
\end{equation}
Here, the subscript $j$ corresponds to the direction. Consequently, the 3-dimensional Gaussian distribution $\phi(\mathbf{x})$ is expressed by Equation \ref{eq:multi_var_normal_distribution}:

\begin{equation}
\phi(\mathbf{x}) = \prod_{j=1}^{3} \frac{1}{\sqrt{2\pi\sigma_{j}^{2}}} \exp \left\{-\frac{1}{2 \sigma_{j}^{2}} \left(x_{j} - \mu_{j}\right)^2 \right\}.
\label{eq:multi_var_normal_distribution}
\end{equation}
Finally, the neutron source $u(E, \mathbf{x})$ is defined by multiplying a energy distribution $N(E)$ at a single energy level $E$ by the distribution obtained in Equation \ref{eq:multi_var_normal_distribution}:

\begin{equation}
u(E, \mathbf{x}) = N(E) \phi(\mathbf{x}) = N(E) \prod_{j=1}^{3} \frac{1}{\sqrt{2\pi\sigma_{j}^{2}}} \exp \left\{-\frac{1}{2 \sigma_{j}^{2}} \left(x_{j} - \mu_{j}\right)^2 \right\},
\label{eq:neutron_source}
\end{equation}
where $\mu_{j}$ represents the component of the mean vector $\mathbf{\mu}$. 
\end{comment}

For every simulation, the single neutron energy $E$ and mean position $\mu_{2}$ are randomly generated within the ranges of $E$ and $y$ are defined by $E \in (0,1)$ in MeV and $y \in [-9, 9]$ in cm, respectively. These values are then employed in Equation \ref{eq:neutron_source} to prepare the neutron source. It should be noted that the variances $\sigma_{1}^{2} = \sigma_{2}^{2} = 1$, $\sigma_{3}^{2} = 0$, and mean values $\mu_{1} = \mu_{3} = 0$ remain fixed throughout the process. The possible regions of the mean position $y$ are indicated by the red boxed area in Figure \ref{fig:maze_input_function} (a). Figure \ref{fig:maze_input_function} (b) shows an example of 15 randomly generated patterns of neutron sources projected onto the y-plane, which confirms that the peak value and peak position are each a random combination.

\begin{figure}[!htbp]
    \centering
    \includegraphics[width=\textwidth]{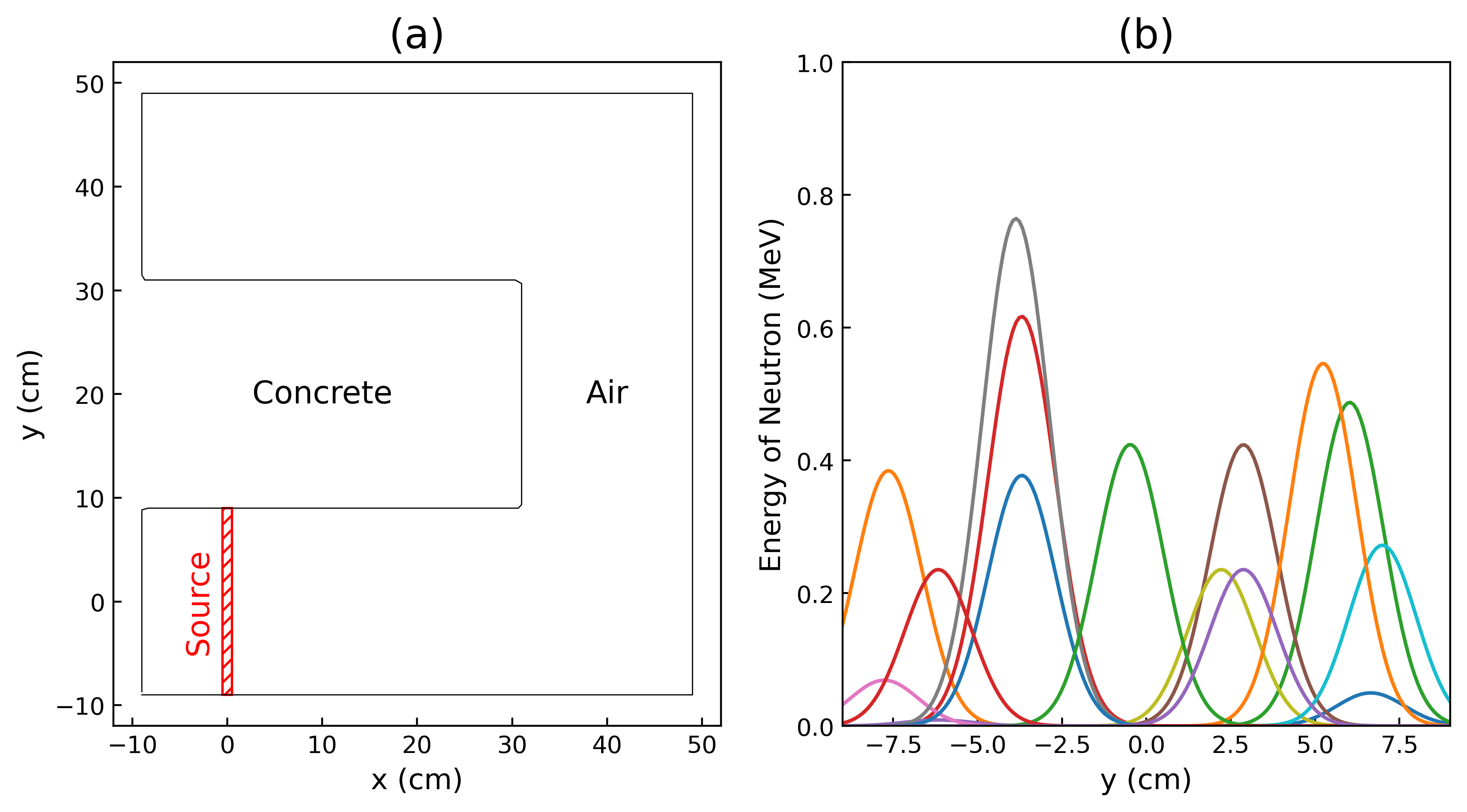}
    \caption{(a) the layout of the concrete maze used in the particle transport simulation. The maze is enclosed by concrete walls, and the interior is filled with air. The neutron source position $y$ is randomly selected, while the $x$ location is fixed at $x=0$. (b) example of randomly generated distribution neutron sources (neutron energy density). The peak value and the Gaussian center position in the y-axis are randomly generated in each simulation run (see Section \ref{sec:source}).}
    \label{fig:maze_input_function}
\end{figure}

\subsubsection{Tally}
\label{sec:tally}
The flux within the geometry is calculated using the [T-Track] tally. To cover potential peak energies up to $1$ MeV, the energy mesh range is defined from $1$ keV to $10$ MeV. For spatial resolution, the geometry is divided into $80$ divisions in both the x- and y-directions, ranging from $-12$ cm to $52$ cm, while a single division is used in the z-direction, covering a range from $-10$ cm to $10$ cm. This results in the definition of $6,400$ sub-regions on the xy-plane, and the neutron flux values are computed for each of these regions. In this calculation, a normalization factor of $1,000$ is set to avoid the default per-particle normalization in PHITS, which would result in a significantly smaller value.

\subsection{Executing the simulations}
\label{sec:simulation_speed}
A total of $1,900$ random combinations of energy $E$ and the mean value of $y$ positions are generated, and simulations are carried out for each combination. Each simulation includes $10^{5}$ neutrons per batch, with a total of $10^{2}$ batches per combination. This setup ensures that the relative error of the neutron flux remains below $10\%$. 

In order to provide a comprehensive comparison between the simulation and surrogate model runtimes, it is important to include information about the computational environment. The simulations are performed on a machine running the Ubuntu 22.04 operating system, equipped with an AMD Ryzen9 3900X CPU (12 Cores/24 Threads) and 64GB DRAM (3,200 MHz). PHITS version 3.24, compiled by our group with the Intel Fortran Compiler (Intel(R) Parallel Studio XE 2020 Update 1 for Linux), is used for hybrid MPI and OpenMP parallel computing. The number of threads (or cores) to be used for OpenMP is set to $2$ threads per core, and one for MPI is $11$ cores. Based on the log data, the time required per simulation was $30.54 \pm 3.76$ seconds. 

%%%%%%%%%%%%%%%%%%%%%%%%%%%%%%%%%%%%%%%%%%%%%%%%%%%%%%%%%%%%
\subsection{Preprocessing}
\label{sec:processing}
A systematic multi-stage protocol is implemented for preprocessing the data to trainining and test the DeepONet model.

Initially, an ensemble of $n$ functions ${u^{1}, u^{2}, \ldots, u^{n}}$ is generated, each representing a hypothetical neutron source distribution scenario within our study. These functions are discretized by sampling them at $m$ uniformly distributed spatial coordinates across the domain of interest. This process is graphically represented in Fig. \ref{fig:flow} (a), providing a visual understanding of the discretization methodology. This study evaluates each function at $m=190$ points along the $y$-axis, maintaining a constant interval width of $0.095$ cm between points to ensure uniform domain coverage. Following the discretization, these functions are partitioned into training and test datasets in an 8:2 ratio. This data division is designed to optimize the training process while comprehensively evaluating the model's predictive capability on unseen data.

As Fig. \ref{fig:flow} (b) represents, the discretized functional data are subsequently used as inputs for particle transport simulations conducted via PHITS on an $80 \times 80$ Cartesian grid. This simulation step produces a matrix dataset of $6,400$ grid point coordinates $(x, y)$, each paired with its corresponding simulated neutron flux values $\psi(x, y)$. This process effectively transforms the abstract functional forms into concrete, quantifiable simulation outcomes. 

Furthermore, as demonstrated in Fig. \ref{fig:flow} (c), while the entire set of $6,400$ $(x, y)$ coordinate pairs with corresponding simulation results $\psi(x, y)$ is available for surrogate model construction, we have methodically created multiple sub-datasets, labeled as Set1 through Set5, to evaluate the impact of data volume on DeepONet's performance. Each dataset is generated by randomly sampling a specific proportion of the total data in increments of 10\%, starting from 50\% and extending up to 90\% of the 6,400 pairs. This stratified sampling strategy facilitates a comprehensive evaluation, allowing us to methodically analyze how the variation in the volume of training data affects the model's predictive accuracy. Creating these subsets, Set1 to Set5 enables a systematic investigation into the relationship between the quantity of training data and the fidelity of the DeepONet model.

\begin{figure}[!htbp]
    \centering
    \includegraphics[width=\textwidth]{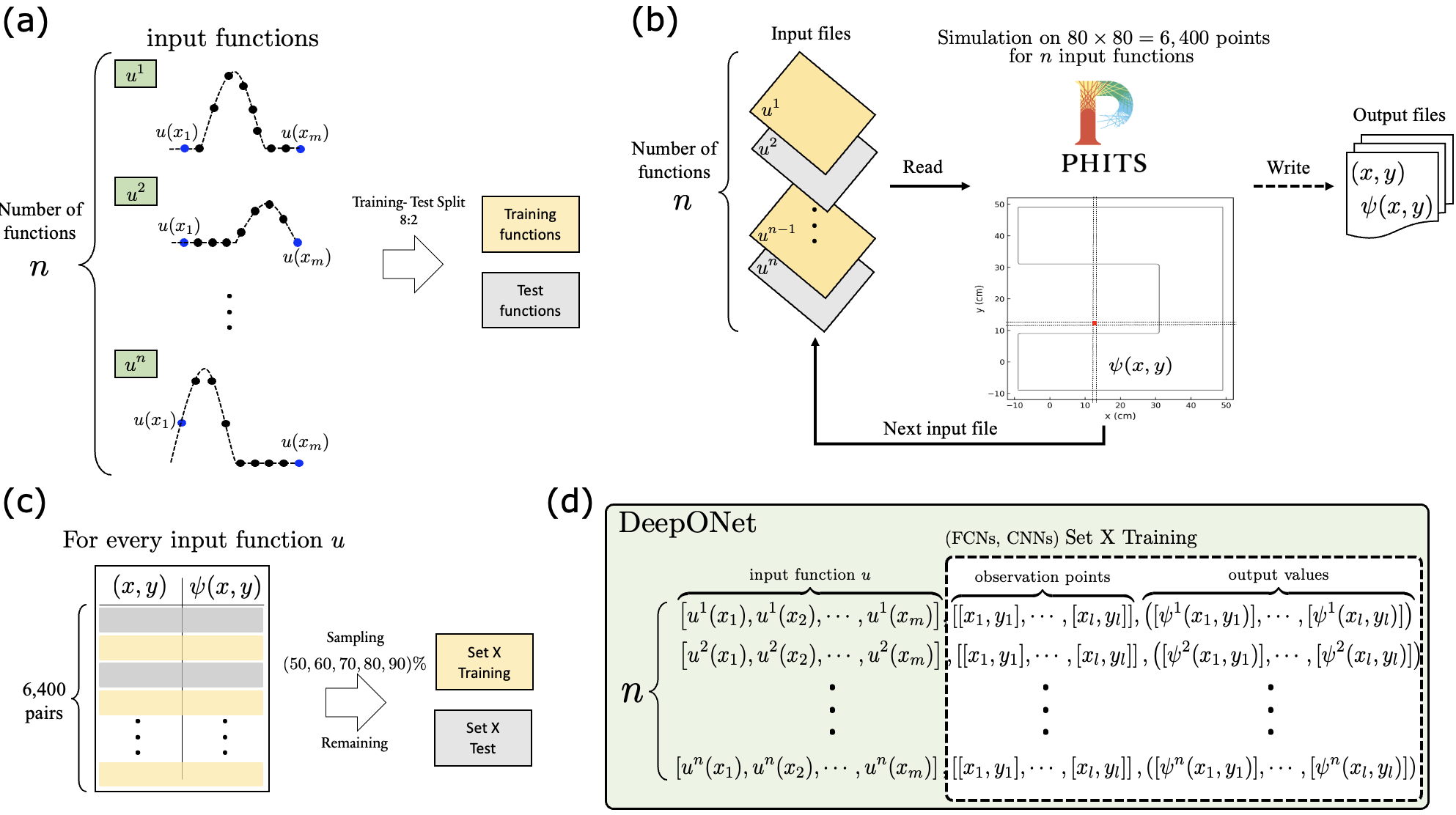}
    \caption{Overview of the data preparation and simulation pipeline for DeepONet and conventional neural network models: (a) Input functions $u^i$ are sampled at $m$ points, with the total number of functions being $n$. These functions are then split into training and test sets using an 8:2 ratio. (b) Each input function is read and processed through a PHITS simulation on a grid of 80 x 80 points to generate output files containing values of $\psi(x, y)$ at corresponding coordinates. (c) The simulation results for every input function $u$ form a dataset comprising 64,000 $(x, y)$ pairs, which are then sampled into subsets for training and testing purposes. (d) The DeepONet framework utilizes the input functions, observation points, and output values for model training, distinguishing it from the training data structure used in FCNs and CNNs, which are represented in a more traditional feature-target format. The subscripts $l$ in the observation points and output values equal to the length of sampled dataset X in panel (c).}
    \label{fig:flow}
\end{figure}

%%%%%%%%%%%%%%%%%%%%%%%%%%%%%%%%%%%%%%%%%%%%%%%%%%%%%%%%%%%%
\subsection{DeepONet Model}
The desired function of a DeepONet model is to obtain the operator that maps between the distribution of neutron source and the 2-dimensional spatial distribution of neutron flux in the maze. In this case, the operator $G$ can be represented as $G: u(E, \mathbf{y}) \mapsto \psi(\mathbf{y})$, where $\mathbf{y} \in \mathbb{R}^{2}$ denotes the position vector in the x-y plane. DeepONet network architecture, model training, and evaluation methods are provided in this section to build the model. The implementation of DeepONet is done using the scientific machine learning library DeepXDE \cite{lu2021deepxde}.

DeepONet utilizes fully connected neural networks for both the branch and trunk networks. The branch network has a layer size of [190, 80, 80], while the trunk network has a layer size of [2, 80, 80] to accommodate the two-dimensional input $(x, y)$.

The Adam optimization algorithm is employed as the optimization method during the training process. The mean $L^2$ relative error is the evaluation metric to assess the model's performance. The model is trained for 10,000 iterations with a learning rate of 0.001. The learning rate determines the step size during gradient descent and influences the convergence speed and accuracy of the training process. These choices of optimization algorithm, evaluation metric, and learning rate are consistent across all training datasets, ensuring a fair and comparable evaluation of the models.

In order to evaluate the performance of DeepONet models trained on each training dataset, four indices are employed: the $\rm R^2$ score, root-mean-squared error (RMSE), mean absolute error (MAE), and the ratio of RMSE to MAE (RMSE/MAE). These indices provide insights into the trained models' accuracy, precision, and general performance. Please refer to Supplementary Material D. %\ref{append:metrics} for details.

%%%%%%%%%%%%%%%%%%%%%%%%%%%%%%%%%%%%%%%%%%

\section{Results \& Discussions}
\subsection{Overall performance of the DeepONet models}
\label{sec:overall}

The evaluation of DeepONet models using different datasets demonstrated their performance on the test dataset in terms of $\rm R^2$ score, RMSE, MAE, and RMSE/MAE, with Table \ref{tab:metrics_all_datasets} presenting the mean and standard deviation of each metric. The results revealed that larger training datasets led to improved model performance. Even with the smallest dataset, the DeepONet models achieved impressive results, with $\rm R^2$ values reaching 0.99 and RMSE and MAE staying within 10\%. For more detailed metrics for all model test data, please refer to Supplementary Material B. %\ref{sec:overall_append}.

Ideally, a well-constructed model should capture the general trends in the data, with only random noise following a normal distribution being represented as errors. In such cases, the ratio of RMSE to MAE should be close to 1.253. However, the observed ratio of approximately 1.53 in all models indicates the presence of data points significantly deviating from the model predictions, suggesting the influence of outliers.

To address this issue, two potential solutions are considered. Firstly, the removal of outliers from the training dataset can improve model generalization and performance by focusing on representative data. Secondly, hyperparameter tuning allows the DeepONet model to adapt more effectively to the dataset's complexities, fine-tuning its performance.

Combining both approaches can lead to a more robust and accurate DeepONet model, enabling reliable predictions even in the presence of challenging data points. Implementing these improvements will enhance the model's effectiveness and applicability, making it a valuable tool in the context of this study.

\begin{table}[htbp!]
\centering
\caption{Performance of DeepONet models evaluated with $\rm R^2$, RMSE, MAE, and RMSE/MAE. The details of datasets are described in Section \ref{sec:processing}.}
\label{tab:metrics_all_datasets}
\begin{adjustbox}{width=\textwidth}
\begin{tabular}{@{}lcccc@{}}
\toprule
\multicolumn{1}{c}{\multirow{2}{*}{Dataset}} & \multicolumn{4}{c}{Metrics}                                                                                \\ \cmidrule(l){2-5} 
\multicolumn{1}{c}{}                            & \multicolumn{1}{c}{$\rm R^2 (\times 10^{-1})$} & \multicolumn{1}{c}{RMSE $(\times 10^{-2})$} & \multicolumn{1}{c}{MAE $(\times 10^{-2})$} & \multicolumn{1}{c}{RMSE/MAE} \\ \midrule
Set1 (50\%)  & $9.93 \pm 0.02$ & $6.61 \pm 1.49$ & $4.27 \pm 1.20$ & $1.56 \pm 0.13$ \\
Set2 (60\%)  & $9.93 \pm 0.02$ & $6.57 \pm 1.47$ & $4.47 \pm 1.22$ & $1.49 \pm 0.10$ \\
Set3 (70\%)  & $9.93 \pm 0.02$ & $6.43 \pm 1.35$ & $4.17 \pm 1.15$ & $1.56 \pm 0.12$ \\
Set4 (80\%)  & $9.95 \pm 0.02$ & $5.71 \pm 1.40$ & $3.84 \pm 1.20$ & $1.51 \pm 0.12$ \\
Set5 (90\%)  & $9.96 \pm 0.02$ & $5.14 \pm 1.33$ & $3.45 \pm 1.16$ & $1.52 \pm 0.12$ \\ 
\bottomrule
\end{tabular}
\end{adjustbox}
\end{table}

\subsection{Comparisons with FCN and CNN}
\label{sec:comparison}
This section compares the model's performance trained with Set1 on the test data against FCN and CNN. As mentioned earlier, DeepONet takes functions as inputs, and the model test evaluates its response to unseen input functions. In this study, 380 test input functions were provided to the model, and the obtained model outputs were compared to the true values. The same metrics used in the previous section were calculated for each input function.

To distinguish each input function, identification numbers (Test ID) were assigned from 1 to 380. Notably, when Test ID 23 and 18 were used, the DeepONet model demonstrated the highest and lowest $\rm R^2$ values, respectively. To compare with conventional ML methods, FCNs and CNNs were trained on these two cases, and the metrics were computed similarly. For more detailed information on the network architectures of FCNs and CNNs in these cases and the parameters used during training, please refer to Supplementary Material C. %\ref{append:compare}.

The calculation results are summarized in Table \ref{tab:benchmark}. In both cases of Test IDs 23 and 18, the DeepONet model outperformed FCN and CNN in terms of $\rm R^2$ score, RMSE, and MAE. Particularly, for Test ID 23, the DeepONet model exhibited significantly better performance with RMSE and MAE values superior by order of magnitude compared to the other models. Additionally, in the challenging case of Test ID 18, where FCN and CNN struggled to build accurate models, DeepONet achieved an impressive $\rm R^2$ value exceeding 0.9, while its RMSE and MAE demonstrated several times better performance than the other models. However, it is noteworthy that the ratio of RMSE to MAE for DeepONet was relatively higher than that of FCN and CNN when comparing the models.

\begin{table}[!htbp]
\centering
\caption{Performances of DeepONet model (Set1), FCN, and CNN. }
\label{tab:benchmark}
\begin{adjustbox}{width=\textwidth}
\begin{tabular}{@{}llcccc@{}}
\toprule
\multirow{2}{*}{Test ID} & \multirow{2}{*}{Model} & \multicolumn{4}{c}{Metrics}               \\ \cmidrule(l){3-6} 
                         &                        & $\rm R^2 (\times 10^{-1})$       & RMSE ($\times 10^{-2}$)    & MAE ($\times 10^{-2}$)      & RMSE/MAE \\ \midrule
\multirow{3}{*}{Highest (ID: 23)} & DeepONet               & 9.97 & 4.77 & 3.06 & 1.56 \\
                         & FCN                 & 9.01 & 25.41 & 20.31 & 1.25 \\
                         & CNN                & 9.10 & 24.68 & 19.00 & 1.30 \\ \midrule
\multirow{3}{*}{Lowest (ID: 18)}  & DeepONet               & 9.79 & 17.42 & 12.95 & 1.35 \\
                         & FCN                   & 5.07 & 86.16 & 73.26 & 1.18
                         
                         \\
                         & CNN                  & 5.10 & 85.92 & 73.47 & 1.17 \\ \bottomrule
\end{tabular}
\end{adjustbox}
\end{table}

%%%%%%%%%%%%%%%%%%%%%%%
In addition to quantitative model performance considerations, confirming the reproducibility of physical phenomena is crucial. In this study, we modeled the spatial distribution of neutron flux using the neutron source distribution in a two-dimensional space as an input function ($G: u(E, \mathbf{y}) \mapsto \psi(\mathbf{y})$). Since the system under consideration lacks fissile material, the Gaussian center position of the neutron source distribution is expected to have the highest value. Based on this fact, we compared the predictions of DeepONet, FCN, and CNN models.

Figure \ref{fig:predictions_id_23} presents the ground truth from the simulation, along with the predictions of DeepONet, FCN, and CNN for Test ID 23. As expected, the highest values are obtained at the Gaussian center position of the neutron source (Figure \ref{fig:predictions_id_23} (a)). The DeepONet model can reproduce the overall distribution of neutron flux, although the peak value at the Gaussian center position is estimated to be small (Figure \ref{fig:predictions_id_23} (b)). In contrast, FCN and CNN fail to accurately reproduce both the Gaussian center location and the overall neutron distribution prediction (Figures \ref{fig:predictions_id_23} (c) and (d)).

\begin{figure}[!htbp]
    \centering
    \includegraphics[width=\textwidth]{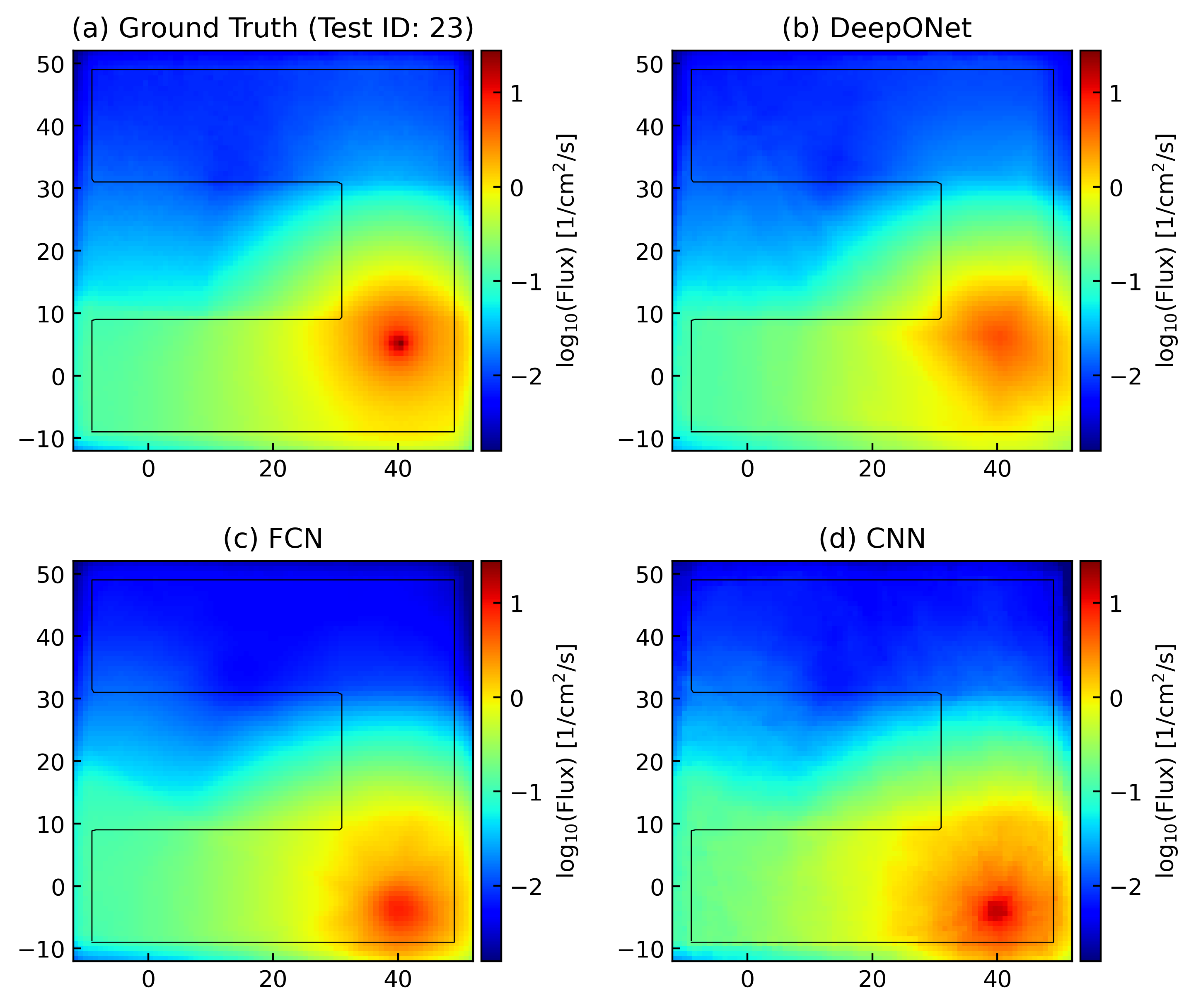}
    \caption{Comparisons between the ground truth, DeepONet, FCN, and CNN predictions for the Test ID of 23.}
    \label{fig:predictions_id_23}
\end{figure}

Figure \ref{fig:predictions_id_18} shows the ground truth from the simulation and the predictions of DeepONet, FCN, and CNN for Test ID 18. The DeepONet model is capable of reproducing the overall distribution of neutron flux, although the Gaussian center position is slightly blurred (Figure \ref{fig:predictions_id_18} (b)). On the other hand, FCN and CNN models do not accurately reproduce the neutron flux distribution, especially in the upper-left regions farthest from the neutron source (Figures \ref{fig:predictions_id_18} (c) and \ref{fig:predictions_id_18} (d)). Notably, in this test case, the neutron flux is overestimated relative to the true value, which could lead to potentially hazardous situations from a radiation protection perspective if underestimated.

Overall, these findings underscore the ability of the DeepONet model better to capture the intricate spatial distribution of neutron flux compared to FCN and CNN models, showcasing its potential for reliable and accurate predictions in nuclear engineering applications.

\begin{figure}[!htbp]
    \centering
    \includegraphics[width=\textwidth]{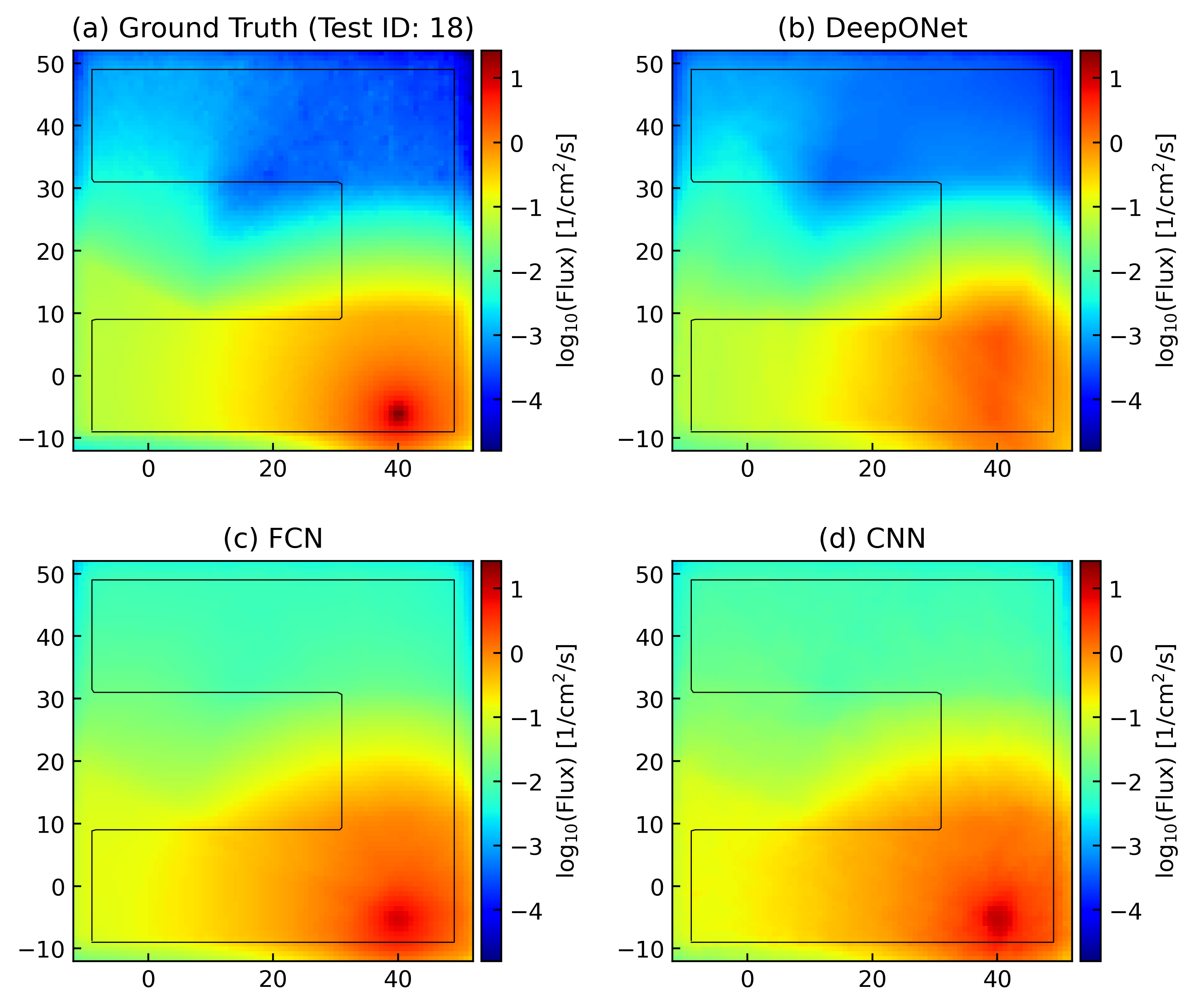}
    \caption{Comparisons between the ground truth, DeepONet, FCN, and CNN predictions for the Test ID of 18.}
    \label{fig:predictions_id_18}
\end{figure}

%%%%%%%%%%%%%%%%%%%%%%%%%%%%%%%%%%%%%%%%%%

%%%%%%%%%%%%%%%%%%%%%%%%%%%%%%%%%%%%%%%%%%
\subsection{Key Discussions}
The study demonstrates the power of DeepONet, which takes functions as input data and constructs operator $G$ in the system using training data. Notably, the prediction accuracy of DeepONet matches or surpasses that of conventional ML methods like FCN and CNN. The use of fixed sensors to extract features from input functions and their integration into the model through a branch network is a compelling concept. Training the model with historical data or high-confidence simulations allows it to handle diverse scenarios, including various accidents in the process.

An essential advantage of DeepONet, shared by many ML-based surrogate models, is its remarkably fast execution speed compared to conventional simulations. While a PHITS simulation took about 30 seconds (see Section \ref{sec:simulation_speed}), DeepONet performed the task in just 0.02 seconds. This remarkable speed makes DeepONet a potent modeling method for digital twin systems, enabling real-time predictions based on data from sensors installed on physical assets.

To further enhance the application of DeepONet, two important issues must be addressed. Firstly, understanding the impact of fixed sensors' number and location on model performance is crucial, considering the limited installation possibilities in harsh environments like nuclear power systems. Optimizing sensor placement and quantity under such constraints will be necessary for accurate and reliable modeling. Secondly, evaluating the DeepONet model requires attention. While overall metrics may indicate excellent performance, certain input functions might produce spurious predictions (see Section \ref{sec:overall} and Section \ref{sec:comparison}). Improving the model evaluation process, including hyperparameter tuning, is necessary to ensure robust and dependable predictions in all scenarios.

By addressing these challenges, DeepONet can be further optimized for digital twin systems, enhancing its potential to predict and analyze current and future systems based on real-time data from physical assets' sensors. This advancement opens new possibilities for various engineering applications, including nuclear engineering and beyond.

%%%%%%%%%%%%%%%%%%%%%%%%%%%%%%%%%%%%%%%%%%
\section{Conclusions}
In conclusion, this research has highlighted the significant potential of DeepONet as a robust surrogate modeling method for digital twin (DT) for nuclear energy systems. The utilization of DeepONet allows for the accurate prediction of complex behaviors and spatial distributions of neutron flux, surpassing the performance of conventional ML methods like FCN and CNN. Its ability to handle functions as input data and construct operator $G$ from training data makes it a valuable tool for capturing the intricate behavior of nuclear systems.

Through extensive benchmarking and evaluation, DeepONet showcased remarkable prediction accuracy and computational efficiency. It demonstrated consistent improvement in performance with increasing training datasets, making it a versatile and scalable solution for various nuclear engineering applications. The speed of execution, significantly faster than traditional simulations, positions DeepONet as a promising method to enable real-time predictions based on sensor data from physical assets.

While DeepONet shows great promise, the study also sheds light on challenges that need further investigation. The impact of fixed sensors and the optimal sensor placement for improved model performance requires careful consideration, given the constraints in harsh operating environments like nuclear power systems. Additionally, developing more effective model evaluation methods is crucial to ensure reliable predictions and robustness.

Overall, this research contributes valuable insights into the advancement of digital twin technology in nuclear engineering. By harnessing the capabilities of DeepONet, researchers and operators can enhance the safety, efficiency, and reliability of nuclear systems, paving the way toward a sustainable and carbon-neutral energy future. As digital twin systems continue to evolve and mature, DeepONet is a promising and innovative approach to revolutionizing nuclear engineering research and applications.

}

%%%%%%%%%%%%%%%%%%%%%%%%%%%%
\bibliographystyle{unsrtnat}
\bibliography{references} 

\end{document}